# LISTA-Transformer Model Based on Sparse Coding and Attention Mechanism and Its Application in Fault Diagnosis

Shuang Liu[1], Lina Zhao[2,*], Tian Wang[3], Huaqing Wang[4]
1, 2, 3, The College of Mathematics and Physics,
Beijing University of Chemical Technology, China
4, The College of mechanical and electrical engineering,
Beijing University of Chemical Technology, China
zhaoln@mail.buct.edu.cn

*Abstract*—**Driven by the continuous development of models such as Multi-Layer Perceptron, Convolutional Neural Network (CNN), and Transformer, deep learning has made breakthrough progress in fields such as computer vision and natural language processing, and has been successfully applied in practical scenarios such as image classification and industrial fault diagnosis. However, existing models still have certain limitations in local feature modeling and global dependency capture. Specifically, CNN is limited by local receptive fields, while Transformer has shortcomings in effectively modeling local structures, and both face challenges of high model complexity and insufficient interpretability. In response to the above issues, this article proposes the following innovative work: A sparse Transformer based on Learnable Iterative Shrinkage Threshold Algorithm (LISTA-Transformer) was designed, which deeply integrates LISTA sparse encoding with visual Transformer to construct a model architecture with adaptive local and global feature collaboration mechanism.
This method utilizes continuous wavelet transform to convert vibration signals into time-frequency maps and inputs them into LISTA-Transformer for more effective feature extraction. On the CWRU dataset, the fault recognition rate of our method reached 98.5%, which is 3.3% higher than traditional methods and exhibits certain superiority over existing Transformer-based approaches.**

*Keywords*—**Visual Transformer, LISTA, Time-frequency diagram, intelligent fault diagnosis of rolling bearings**

## 1 Introduction

As the core component of industrial equipment, the running state of rolling bearing directly determines the performance and safety factor of the equipment. Statistics show that about one-third of the faults of mechanical equipment are caused by the failure of rolling bearings. Therefore, it is of great engineering significance to carry out efficient bearing fault diagnosis research to ensure the stable operation of equipment, reduce maintenance costs and reduce shutdown losses. With the iteration of industrial technology and the vigorous development of Internet and machine learning technology, the importance of bearing fault diagnosis has become increasingly prominent. Neural network-based methods show strong ability in feature extraction and fault recognition, which has brought new opportunities to this field. However, most traditional deep learning methods remain inefficient at extracting long-distance dependent features. Transformer model initially achieved a major breakthrough in the field of natural language processing. With its excellent ability to capture temporal features, it has become an ideal scheme for processing sequential data. After it is introduced into the field of bearing fault diagnosis, it can effectively analyze the long-distance dependence in the vibration signal, and then accurately extract the fault characteristics, providing a new technical path for bearing state recognition under complex working conditions. However, the transformer neural network (VSTNN [1]) based on vibration signals improves the accuracy and efficiency of fault diagnosis through the new vibration signal marking strategy and multi head attention mechanism, but its computational complexity is high, and there is a risk of over fitting under uneven samples.

Therefore, in order to ensure the stable operation of mechanical equipment, reduce maintenance costs and downtime, effective fault diagnosis of rolling bearings is of great significance. With the continuous development of industrial technology, the importance of bearing fault diagnosis has become increasingly prominent in the context of Internet and machine learning. At present, the main research methods of intelligent fault diagnosis are:

The conventional fault diagnosis method based on signal processing is mainly to diagnose the equipment fault by analyzing the characteristics of the signal extracted. These methods have been developed in the field of signal processing for decades, with mature technology and wide applications. Representative methods include envelope analysis, cepstrum analysis, principal component analysis, wavelet transform [2], Fourier transform, empirical mode decomposition [3], support vector machine [4] and blind source separation technology [5].

In the process of bearing compound fault diagnosis based on machine learning, the training set data is used to train the machine learning model. Traditional diagnosis methods mainly rely on artificial experience, signal processing technology and simple machine learning algorithm. However, these methods have great limitations in the case of less sample data, and the generalization ability and diagnostic accuracy of the model are limited. Taking random forest [6] (RF) as an example, machine learning can select different classification algorithms or regression prediction models according to different fault types. In the training process, in order to achieve the optimal performance, the parameters of the model need to be adjusted accordingly. ABBASION and other [7] researchers used wavelet analysis and support vector machine (SVM) to optimize the signal decomposition level and classify various faults of bearings. Yin et al. [8] reviewed the research and development of fault diagnosis based on support vector machine (SVM). Widodo et al. [9] introduced the method of using support vector machine (SVM) to monitor the state of the machine in real time and the fault.

In the field of bearing fault diagnosis, deep learning (DL) has attracted much attention. [10] proposed a new fault diagnosis method based on the sliding window processing and CNN-LSTM model combined with convolutional neural network (CNN) and short-term memory network (LSTM); Chen et al. [11] proposed an automatic learning feature neural network using one-dimensional time sequence signal as input. After obtaining the feature, the short-term and long-term memory network was used for fault classification; Gu et al. [12] proposed a hybrid fault diagnosis method suitable for small sample processing, which combines techniques such as variational mode decomposition (VMD), continuous wavelet transform (CWT) and support vector machine (SVM); Wen et al. [13] proposed a new convolutional neural network for fault diagnosis, which is based on lenet-5 structure. This method transforms one-dimensional time-frequency signal into two-dimensional (2-D) time-frequency image, and extracts the features of the converted two-dimensional image, so as to eliminate the influence of manual features.

Traditional fault diagnosis methods mainly rely on artificial experience, signal processing technology and simple machine learning algorithms, such as support vector machine (SVM). However, when dealing with complex nonlinear fault signals, these methods have certain limitations, especially when the sample data is small, the generalization ability and diagnosis accuracy of the model are limited. In the field of fault diagnosis, in-depth learning technology has been widely used in recent years. Convolutional neural network and cyclic neural network are used to extract fault features, and good results are achieved. However, CNN has limitations in capturing the overall information, and RNN is prone to gradient disappearance or gradient burst when processing long series of data. The emergence of transformer architecture provides a new way to solve these problems, and its self-care mechanism can effectively capture the overall dependence in the data.

In this chapter, a cross modal compression method of vibration signal time-frequency diagram based on LISTA - Transformer is proposed, which reduces the input redundancy in the CWRU data set while retaining the impact characteristics of key faults. The continuous wavelet transform method is used to preprocess the data on the original vibration signal, convert it into a two-dimensional time-frequency image, and input this image into the LISTA transformer model architecture for training. The model combines the advantages of sparse coding and transformer architecture, and can simultaneously extract the local features and global dependencies in the time-frequency image, so as to improve the performance of fault diagnosis.

The rest of this paper is organized as follows: Section 2 will introduce the prior knowledge; Section 3 details our approach; Section 4 will show the results and analysis of bearing fault diagnosis experiment; Finally, section 5 summarizes this chapter.

## 2 Prior knowledge

### 2.1 Time frequency map transformation method

Signal processing is an important field in modern science and engineering. The time domain information of signal represents the relationship between signal amplitude and time, and the frequency domain information of signal reveals the relationship between signal frequency and energy. When dealing with non-stationary signals, the traditional time-domain and frequency-domain analysis methods are often insufficient to provide sufficient information. By providing time and frequency information at the same time, time-frequency analysis method can better deal with such signals. Transforming signals into time-frequency maps is an important task in signal processing, which can help us better understand the changes of signals in time and frequency domain. The following describes four methods for converting the original signal into time-frequency map: short time Fourier transform (STFT), continuous wavelet transform (CWT), Wigner Ville distribution (Wigner Ville), Hilbert Huang transform (HHT).

A segment of fault data at the drive end is randomly selected from the CWRU rolling bearing dataset. The following is the time domain diagram, frequency domain diagram and the time-frequency diagram converted by four methods:

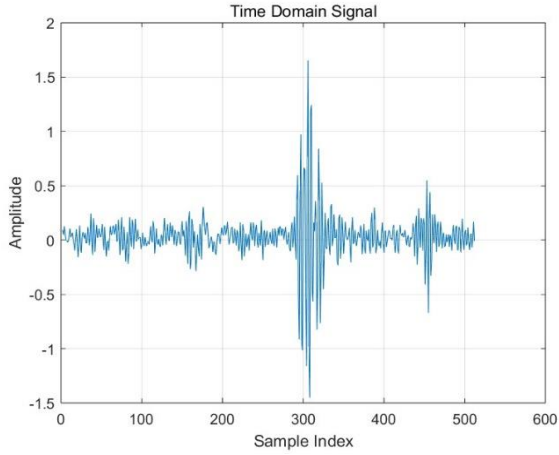

**Figure 1** Time domain plots of a randomly selected segment of drive end fault data in the rolling bearing dataset of Case Western Reserve University (CWRU)

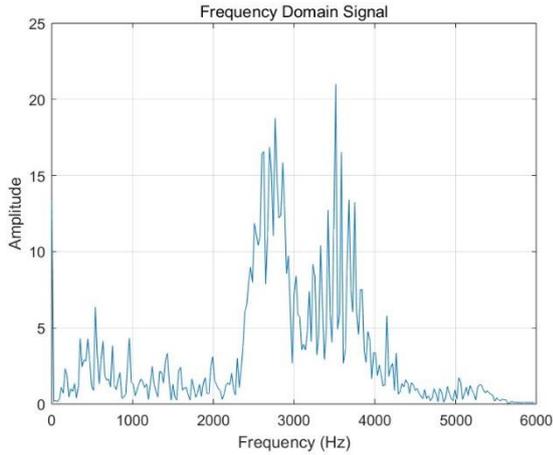

**Figure 2** Frequency domain plots of a randomly selected segment of drive end fault data in the rolling bearing dataset of Case Western Reserve University (CWRU)

(1) STFT

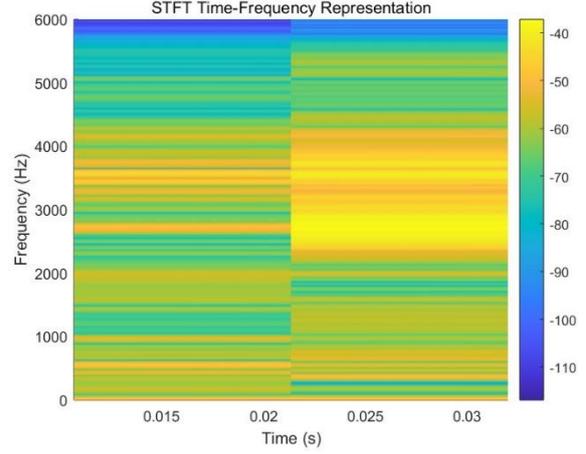

**Figure 3** STFT time-frequency plot of a randomly selected segment of drive end fault data in the CWRU rolling bearing dataset

For stationary signals, Fourier transform (FT) is an important method of analysis and processing. However, for the non-stationary signals widely existing in nature, Fourier transform can only obtain the global spectrum information, while the frequency change information is lost in local time, so it can not accurately characterize the characteristics of non-stationary signals. The Fourier transform of continuous time signal is defined as follows:

$$F(\omega) = \int_{-\infty}^{\infty} f(t) e^{-j\omega t} dt \qquad (1)$$

Where, $F(\omega)$ is the frequency domain representation, $\omega$ is the angular frequency.

STFT of continuous time signal is defined as follows:

$$STFT\{f(\tau)\}(t,\omega) = \int_{-\infty}^{\infty} f(\tau) \omega(\tau-t) e^{-j\omega\tau} d\tau \qquad (2)$$

Where: $f(t)$ is the one-dimensional time domain signal to be analyzed; $\omega(t)$ is a window function used to limit the time range of analysis; $t$ is a time variable, indicating the center position of the window function; $\omega$ is the angular frequency, indicating the frequency component of the signal. When the central time $t$ is selected, the window function is $h(\tau - t)$, and the short-time Fourier transform acts on the window function.

(2) CWT

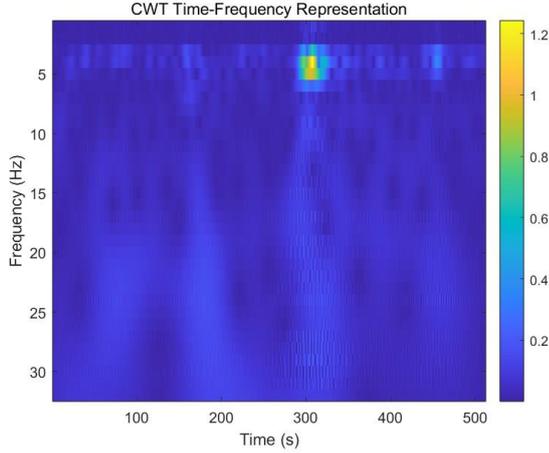

**Figure 4** CWT time-frequency plot of a randomly selected segment of drive end fault data in the CWRU rolling bearing dataset

Continuous wavelet transform (CWT) inherits and develops the idea of local segmentation adopted by short-time Fourier for non-stationary signal processing. At the same time, it overcomes the shortcoming of fixed width of short-time Fourier window function, and provides a time window that changes with frequency for signal time-frequency transformation. The continuous wavelet transform of a signal is defined as the convolution integral of a signal $f(t)$ and a wavelet function $\psi(t)$:

$$CWT_f(a,b) = \int_{-\infty}^{\infty} f(t) \frac{1}{\sqrt{|a|}} \psi^*(\frac{t-b}{a}) dt \quad (3)$$

Where, $a$ is the scale parameter, which controls the broadening or compression of the wavelet function in time and frequency, and the larger $a$ represents the lower frequency (corresponding to the lower time resolution); $b$ is the displacement parameter, indicating the position of wavelet along the time axis. $\psi(t)$ is the mother wavelet, $\psi^*$ is the complex conjugate of wavelet function. $1/\sqrt{|a|}$ is a normalization factor introduced to keep the energy constant.

(3) Wigner-Ville

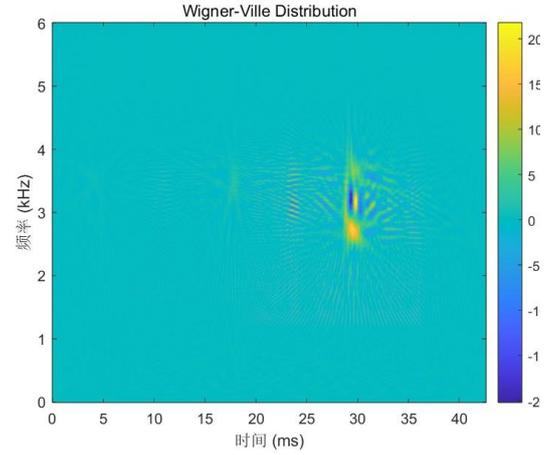

**Figure 5** Wigner-Ville time-frequency plot of a randomly selected segment of drive end fault data in the CWRU rolling bearing dataset

The American physicist Eugene Wigner first proposed the concept of Wigner Ville distribution (WVD) in 1932. Later, Ville applied this distribution to the field of signal processing in 1948. The expression of the instantaneous autocorrelation function of the signal is as follows:

$$r_s(t,\tau) = s(t+\frac{\tau}{2})s^*(t-\frac{\tau}{2}) \quad (4)$$

The definition of WVD is as follows:

$$WVD(t,\omega) = \int_{-\infty}^{\infty} s(t+\frac{\tau}{2})s^*(t-\frac{\tau}{2})e^{-j\omega\tau}d\tau \quad (5)$$

The Wigner Ville distribution of signal has high time resolution and frequency resolution, and has good time-frequency aggregation, which overcomes the shortcomings of STFT and is suitable for analyzing unstable random signals.

(4) HHT

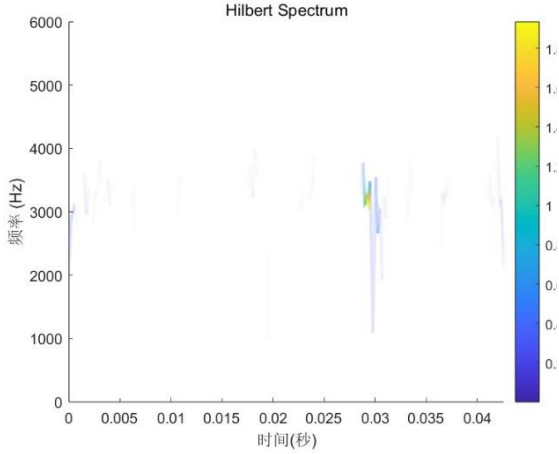

**Figure 6** HHT time-frequency plot of a randomly selected segment of drive end fault data in the CWRU rolling bearing dataset

For the signal $s(t)$ to be analyzed, all local maximum and minimum points are determined first. Then, these maximum points and minimum points are fitted respectively to generate the corresponding maximum envelope and minimum envelope, and the average value of the two envelope is calculated.

$$m_1 = \frac{e_{max}(t) + e_{min}(t)}{2} \quad (6)$$

$$y_1(t) = s(t) - m_1 \quad (7)$$

The signal satisfying the intrinsic mode function condition is the first intrinsic mode function component of the original signal, which is marked as $y_1(t) = c_1(t)$ satisfying the condition, then $c_1(t)$ is the first intrinsic mode function component of the signal $s(t)$.

$$s(t) = \sum_{i=1}^{n} c_i(t) - c_n(t) \quad (8)$$

Perform Hilbert transform on the natural mode function components of each order:

$$H[c(t)] = \frac{1}{\pi} PV \int_{-\infty}^{\infty} \frac{s(\tau)}{t - \tau} d\tau \quad (9)$$

$$\omega(t) = \frac{d[\arctan(\frac{\hat{s}(t)}{s(t)})]}{dt} \quad (10)$$

The original signal can be represented as:

$$s(t) = \mathrm{Re}(\sum_{i=1}^{n} a_i(t) e^{j\phi_i(t)}) = \mathrm{Re}(\sum_{i=1}^{n} a_i(t) e^{j\int \omega_i(t) dt}) \quad (11)$$

After unfolding, it becomes the Hilbert spectrum, which can be written as:

$$H(\omega, t) = \mathrm{Re}(\sum_{i=1}^{n} a_i(t) e^{j\int \omega_i(t) dt}) \quad (12)$$

Hilbert spectrum is a joint spectrum with complete time, frequency, and energy, in which the frequency changes in different time periods and the energy changes with time and frequency can be observed simultaneously.

### 2.2 LISTA

A basic linear inverse problem:

$$X = W_d Z + w \quad (13)$$

where, $X \in R^m$ is the measurement value, $W_d \in R^{m \times n}$ is the measurement matrix and is unknown, $w$ is the unknown noise, and $Z$ is the sparse signal that needs to be reconstructed.

The classic method for solving equation (13) is the least squares (LS) method:

$$Z_{LS} = \arg\min_{Z} \| W_d Z - X \|_2^2 \quad (14)$$

The LASSO problem can be expressed as:

$$Z = \arg\min_{Z} \| W_d Z - X \|_2^2 + \alpha \| Z \|_1 \quad (15)$$

where, $\alpha > 0$ is regularization parameters.

The ISTA algorithm updates Z through a soft threshold operation in each iteration, and its specific iteration format is as follows:

$$Z(k+1) = h_{\alpha/L}(Z(k) - \frac{1}{L}W_d^T(W_d Z(k) - X)) \quad (16)$$

After expanding and merging similar items, the following result is obtained:

$$Z(k+1) = h_\theta(W_e X + S Z(k)) \quad Z(0) = 0 \quad (17)$$

where, Filter matrix: $W_e = \frac{1}{L}W_d^T$,

Mutual inhibition matrix: $S = I - \frac{1}{L}W_d^T W_d$,

Soft threshold shrinkage function:

$$[h_\theta(V)]_i = sign(V_i)(|V_i| - \theta_i)_+$$

The block diagram of ISTA method is as follows:

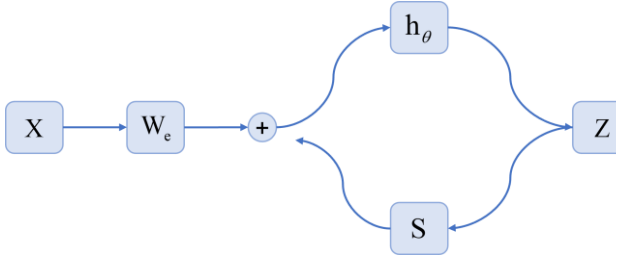

**Figure 7** Block diagram of the ISTA algorithm for sparse coding

LISTA is typically implemented as a recurrent neural network (RNN) with a single hidden layer. The network takes the current estimate and measurement matrix of sparse signals as inputs and outputs the next estimate. It learns all the sums in the process and then uses these parameters to solve sparse encoding. As it is a learning algorithm, it naturally involves the processes of forward reasoning and backpropagation. The architecture of the encoder will be represented as, where represents all trainable parameters in the encoder. The training encoder uses stochastic gradient descent to minimize the loss function. The definition of the loss function is:

$$L(W) = \frac{1}{P}\sum_{p=0}^{(P-1)} L(W, X^P) \text{ with}$$

$$L(W, X^P) = \frac{1}{2}\| Z^{*P} - f_e(W, X^P) \|^2 \quad (18)$$

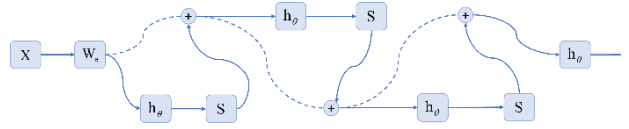

**Figure 8** Block diagram of the LISTA algorithm for sparse coding

## 2.3 Vision Transformer

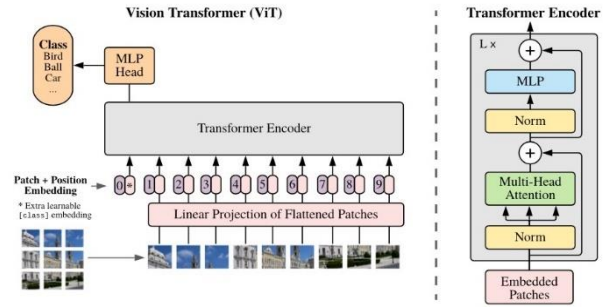

**Figure 9** Model overview of ViT

**Patch Embedding** As one of the core steps of Vit, this process realizes the transformation of image from two-dimensional representation to one-dimensional sequence representation. The operation of embedding image blocks is essentially the process of mapping each image block to a vector representation, so as to complete the above key serialization conversion. Each image block is embedded through a linear layer, which maps it from the pixel space to a higher dimensional feature space. This embedding process not only provides rich feature representation for the model, but also retains the spatial structure information of each image block.

$$z_0 = [x_{class}; x_p^1 E; x_p^2 E; L\ ; x_p^N E] + E_{pos} \quad (19)$$

Where, $x \in R^{H \times W \times C}$ is the original image, $x_p \in R^{N \times (P^2 \cdot C)}$ is a 2D image block sequence, $(H, W)$ is the resolution of the original image, C is the number of channels of the original image, P is the size of the image block.

**Transformer Encoder**

$$[q, k, v] = zU_{qkv} \quad U_{qkv} \in \mathbb{R}^{D \times 3D_h}$$
$$A = soft\,max(qk^T/\sqrt{d_k}) \quad A \in \mathbb{R}^{N \times N}$$
$$SA(z) = Av$$
$$MSA(z) = [SA_1(z); SA_2(z); \cdots; SA_k(z)]U_{msa} \quad (20)$$
$$U_{msa} \in \mathbb{R}^{k \cdot D_h \times D}$$
$$z'_l = MSA(LN(z_{l-1})) + z_{l-1}$$
$$z_l = MLP(LN(z'_l)) + z'_l$$

where $q, k, v$ is the query vector with dimension $d_k$, the key vector and a dimension is the $d_v$ value vector. These three vectors are created by multiplying the word embedding and the weight matrix $U_{qkv}$. The multi head attention function combines the outputs of k attention mechanisms and multiplies them with the projection matrix $U_{msa}$ to obtain the output of the multi head attention mechanism.

## 3. Sparse Transformer Based on LISTA

Local features and global representation in computer vision and machine learning are two complementary methods for understanding and processing images or data. Local features focus on feature extraction of information contained in small regions or patches in the image; On the other hand, the global representation aims to capture the overall structure and content of the image. In many applications, it is beneficial to combine local features with global representation to take advantage of these two methods. For example, in object recognition tasks, local features can help identify specific attributes of objects, while global representation provides context and overall scene understanding. Deep learning models, especially those using convolution layer and full connection layer, are particularly good at this combination because they can capture local patterns and integrate them into a comprehensive global representation. This synergy allows a more robust and accurate model that can handle changes in scale, direction, and partial occlusion while still understanding the broad background of the image.

The deep integration of LISTA and Vit aims to combine their advantages to improve the performance of image processing tasks. Lista focuses on sparse signal recovery and local feature extraction, while vit performs well in capturing the global dependencies of images. The combination of these two methods can give full play to their respective advantages, so as to improve the local and global characteristics of the model. In LISTA transformer, we feed the global context of transformer block into LISTA to enhance the global awareness of LISTA branches. Similarly, the local features of the LISTA block are gradually input into the transformer block to enrich the local details of the transformer branch. Such a process constitutes interaction.

Lista is sparse in nature. By using the iterative threshold shrinkage method, in the learning process of LISTA, the model can automatically select some important features or connections according to the threshold, and make the weight of unimportant features or connections zero or very small, so that the model can automatically select important features or connections in the learning process, and will be or close to zero. Adding LISTA to the back of the self attention mechanism block will sparse constrain and optimize the attention weights generated by the self attention mechanism, and only the attention weights between key positions will be retained, making the attention weight matrix sparse.

Although the self attention mechanism can capture the global dependencies in the sequence, its attention weight matrix may be dense. However, LISTA can filter and sparse the weight matrix, so that each step of the model only focuses on a small amount of important information, so the model can focus more on the main part during feature extraction and information transmission, increasing the sparsity of the model; On the other hand, because the sparsity of the model is improved, the operation of the model is more concise, which will also save part of the calculation.

Lista has a clear mathematical theoretical basis. And its core idea is to solve the sparse coding problem through iterative optimization. Each iteration is the product of the weight matrix and the input data, and then the nonlinear threshold contraction function is applied. These operations have a clear definition and explanation in mathematics. For example, the weight matrix can be regarded as the basis vector in dictionary learning, and the threshold shrinkage function realizes the sparse processing of features. When LISTA is integrated into the DeiT model, its mathematical composed of LISTA branch and transformer branch, as shown in Figure 10. This parallel structure represents that the two branches can retain local features and global representation to the greatest extent.

In the training process, the cross entropy loss function is used to supervise the classifier.

Transformer branch: this branch cascades N transformer blocks after image block embedding and position embedding.

Given the input, the forward process of the 1st block in the principle is still applicable, which makes the whole model training and reasoning process have a clearer mathematical transformer branch can be expressed as equations (20).

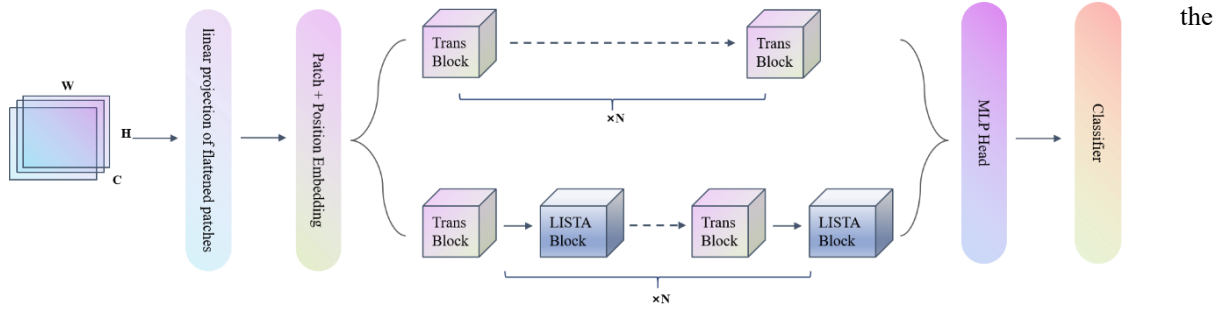

**Figure 10** Model architecture: add LISTA blocks to the network framework after each Transformer Block

LISTA branch: this branch cascades N LISTA

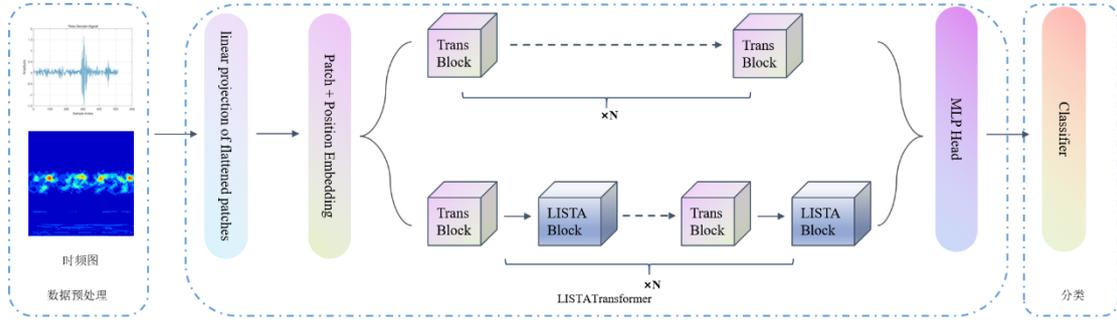

**Figure 11** Structure of model based on time-frequency diagram and LISTA-Transformer

explanation. This combination can be seen mathematically as the constraint optimization of the original self attention mechanism. By sparsely constraining the weights, the model can more effectively focus on the key information, reduce unnecessary calculation and interference, and make the distribution of attention weights more reasonable and interpretable.

Specifically, LISTA transformer is composed of backbone module, two branches, weighted average connecting two branches, and classifier. Lista module, a 7-layer iterative layer, extracts the initial local features (such as edge and texture information), and then provides the features to two branches. The overall architecture is transformer blocks after image block embedding and position embedding (one LISTA block is deeply fused after each transformer block).

The forward process of a LISTA block can be expressed as:

$$z_l = LISTA(z_l^{'}) \\ = h_\theta(W_e z^{l'} + S z_{l-1}) \quad (21)$$

MLP blocks can be represented as:

$$z_{l+1} = MLP(\alpha z_l^{'} + \beta z_l) \quad (22)$$

As shown in Figure 11, the overall architecture of rolling bearing fault diagnosis model based on time-frequency diagram and LISTA transformer includes data preprocessing, LISTA transformer architecture and classifier. The specific steps are as follows: firstly, the vibration signal is transformed from one-dimensional time-series data to two-dimensional time-frequency image using continuous wavelet transform method; Secondly, the obtained time-frequency map is flattened through the linear layer and combined with the position coding, and then input into the LISTA transformer model architecture. In this process, the transformer's multi head attention mechanism plays a major role. It can capture the feature vector of more comprehensive fault diagnosis information, and extract the local features of fault diagnosis through the local focus ability of LISTA iteration; Finally, the training results are output through multi-layer perceptron and classification layer.

## 4. Experiment

### 4.1 Experimental Setup

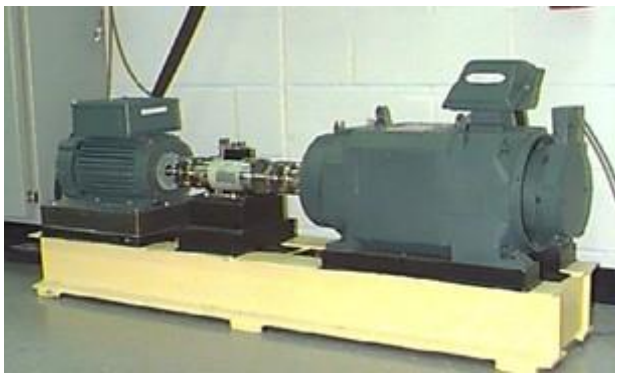

**Figure 12** Bearing Fault Experimental Platform

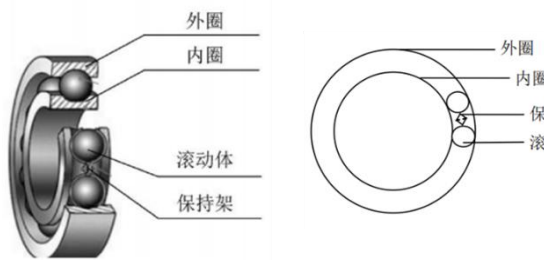

**Figure 13** Bearing structure

The test platform [14] is shown in Figure 12. The CWRU bearing test platform consists of a 1.5KW (2 horsepower) motor (left side of the figure); A torque sensor/decoder (at the middle connection in the figure); A power tester (right side of the figure); Electronic controller (not shown). The bearing structure to be tested is shown in Figure 13.

The damage points of the outer ring of the bearing at the drive end and the fan end were placed at three different positions at 3 o'clock, 6 o'clock and 12 o'clock respectively. It is obtained by collecting vibration signals with 16 channel data recorder. By placing the accelerometer on the drive end of the motor housing, the fan end and the base (normal data), the vibration data of three different positions can be obtained. The difference of these data is that the physical phenomena and characteristics reflected by the measured vibration signals are different, which provides different data for fault diagnosis and monitoring. The power and speed are measured by the torque sensor/decoder. Under the load conditions of 0, 1, 2 and 3 horsepower motors, the machined fault bearings were reinstalled into the test motor, and the vibration acceleration signal data were recorded in the experiment.

Data set the bearing data set of Western Reserve University provides the experimental data under different working conditions, including different speed, load and other parameters. Each failure mode has multiple samples under different working conditions. In this chapter, the accelerometer data of the drive end of the 0.007 inch inner ring fault bearing under normal working conditions and 0 horsepower load are selected as the real data samples for the test.

### 4.2 Data Preprocessing

The data set was divided into training set (70%), validation set (20%) and test set (10%). During each training and testing, the data set is divided randomly.

The continuous wavelet transform is used to obtain the two-dimensional time-frequency diagram. After processing, the time-frequency diagram of the signals of rolling element fault (b), inner ring fault (IR), outer ring fault (or) and normal under the load condition of 0 horsepower is shown in Figure 14.

In the fault diagnosis model based on visual transformer (ViT), the configuration parameters of network architecture directly affect the training efficiency and classification performance of the model. By exploring the quantitative impact of key structural parameters on the performance of the model, this chapter determines the optimal combination of parameters:

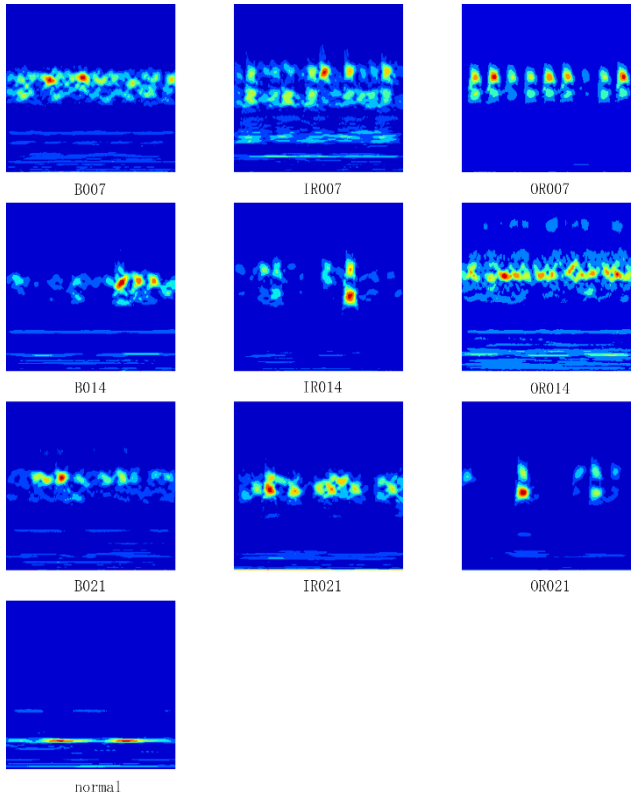

Figure 14 Signal time-frequency diagrams of rolling element fault (B), inner ring fault (IR), outer ring fault (OR), and normal under load condition of 0 horsepower

(1) the trade-off between input resolution and computational efficiency. Improving the size of the input time-frequency map can retain more details in theory, but in the test, it is found that the training time is significantly increased after the resolution exceeds 32 × 32, while the classification accuracy is only improved by less than 0.5%. Therefore, 32 × 32 is selected as the optimal parameter to balance computing resources and diagnostic accuracy. (2) During the parameter configuration of embedded dimension and hidden dimension, it is found that when the embedded dimension is set to 64 and the hidden dimension is set to 128, the accuracy of the test set is the highest and the training time is moderate. (3) When the parameters of transformer coding layer are increased from 2 layers to 6 layers, the accuracy of verification set shows a trend of first increasing and then decreasing. Although the deep network can extract more complex fault features, the over fitting phenomenon intensifies after more than four layers. Finally, two-layer structure is selected to balance the model complexity and generalization performance. (4) Compared with other parameters, the number of attention heads has less effect on the experiment, and the number of attention heads is set to 2.

### 4.3 Result Analysis

In order to evaluate the performance of the proposed model, we conducted detailed experiments on the CWRU dataset. The experimental results show that the proposed fault diagnosis model based on time-frequency map and lista transformer achieves significant fault diagnosis accuracy on the CWRU data set. Compared with the traditional methods based on CNN and SVM and Transformer-based fault diagnosis methods, this model has obvious advantages in diagnosis accuracy and diagnosis time.

Table 1 shows the fault diagnosis accuracy of different models on the CWRU dataset. It can be seen from the table that the accuracy of the proposed model based on time-frequency map and lista transformer on the test set is 98.5%, which is significantly higher than the traditional methods based on CNN and SVM. Transformer-based techniques consistently outperform conventional models, with the lowest-performing transformer (97.8%) exceeding the best traditional method. This validates the effectiveness of attention mechanisms in capturing complex temporal patterns in vibration signals. Also, our model's 0.7% improvement over the baseline transformer indicates that the proposed

LISTA mechanism effectively strengthens temporal feature extraction.

## 5. Conclusion and Future Work

In this paper, the sparse transformer based on the learnable iterative threshold shrinkage algorithm is applied to the rolling bearing fault diagnosis, and a rolling bearing fault diagnosis model based on time-frequency map and LISTA-Transformer is proposed. By combining the advantages of time-frequency analysis and deep learning, the characteristic information in the rolling bearing vibration signal can be effectively extracted, and the cross modal diagnosis framework is further constructed to improve the performance of fault diagnosis.

**Table 1** The accuracy of fault diagnosis using different methods on the CWRU dataset

|  | Method | Accuracy（%） |
| --- | --- | --- |
| Traditional Diagnostic Methods | SVM | 95.2 |
|  | CNN | 96.8 |
|  | LSTM | 97.2 |
|  | Spark-RFA[15] | 96.51 |
|  | Spark-IRFA[15] | 97.22 |
| Transformer-based Fault Diagnosis Methods | 1D-ViT-1Depth[16] | 98.07 |
|  | Swin Transformer-GAMCNN | 97.54 |
|  | Swin Transformer-ResNet | 98.45 |
|  | SSA-SL Transformer[17] | 95.54 |
|  | Transformer | 97.8 |
|  | LISTA-Transformer（ours） | **98.5** |

**Acknowledgements** This work is supported by the National Natural Science Foundation of China (Research on general time series data generation method for industrial Internet, Project No. 92367111).

# About the authors

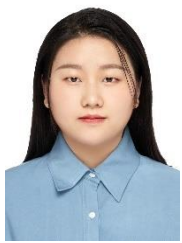

Shuang Liu received her master's degree from Beijing University of chemical technology. Her research focuses on deep learning, sparse low rank model and bearing fault diagnosis.

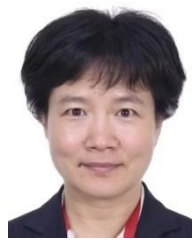

Lina Zhao received the Ph.D. degree in applied mathematics from the Chinese Academy of Sciences, Beijing, China, in 2004. She is currently a Professor with the Beijing University of Chemical Technology, Beijing, China. Her research interests include computational and applied mathematics, which includes geometric problems in computer vision, image processing, and information mining from 3-D Big

Data.(Email:zhaoln@mail.buct.edu.cn)